\documentclass[manuscript, nonacm]{acmart}
\acmConference[]{}{}{}{} 
\renewcommand\footnotetextcopyrightpermission[1]{} 
\usepackage{adjustbox}
\usepackage{caption}
\usepackage{tcolorbox}
\usepackage{multirow}
\usepackage{listings}
\usepackage{xcolor}
\usepackage{fancyvrb}
\usepackage{enumitem}
\usepackage{float}
\usepackage{comment}
\usepackage{titlesec}
\titlespacing*{\section}
{0pt}{1.5ex plus 1ex minus .2ex}{1.3ex plus .2ex}
\titlespacing*{\subsection}
{0pt}{1.5ex plus 1ex minus .2ex}{1ex plus .2ex}
\titlespacing*{\subsubsection}
{0pt}{1.5ex plus 1ex minus .2ex}{1ex plus .2ex}
\definecolor{lightgray}{RGB}{245,245,245}
\definecolor{darkgray}{RGB}{100,100,100}

\AtBeginDocument{%
  }

\begin{document}

\title{The Optimization Paradox in Clinical AI Multi-Agent Systems}


\author{Suhana Bedi}
\email{suhana@stanford.edu}
\affiliation{%
  \institution{Stanford School of Medicine}
  \city{Stanford}
  \state{CA}
  \country{USA}
}

\author{Iddah Mlauzi}
\email{iddah@stanford.edu}
\affiliation{%
  \institution{Department of Computer Science, Stanford University}
  \city{Stanford}
  \state{CA}
  \country{USA}
}

\author{Daniel Shin}
\email{dshin@stanford.edu}
\affiliation{%
  \institution{Department of Computer Science, Stanford University}
  \city{Stanford}
  \state{CA}
  \country{USA}
}

\author{Sanmi Koyejo}
\email{sanmi@stanford.edu}
\affiliation{%
  \institution{Department of Computer Science, Stanford University}
  \city{Stanford}
  \state{CA}
  \country{USA}
}

\author{Nigam H. Shah}
\email{nigam@stanford.edu}
\affiliation{%
  \institution{Department of Medicine, Stanford School of Medicine}
  \city{Stanford}
  \state{CA}
  \country{USA}
}

\renewcommand{\shortauthors}{Bedi et al.}

\begin{abstract}
  Multi-agent artificial intelligence systems are increasingly deployed in clinical settings, yet the relationship between component-level optimization and system-wide performance remains poorly understood. We evaluated this relationship using 2,400 real patient cases from the MIMIC-CDM dataset across four abdominal pathologies (appendicitis, pancreatitis, cholecystitis, diverticulitis), decomposing clinical diagnosis into information gathering, interpretation, and differential diagnosis. We evaluated single agent systems (one model performing all tasks) against multi-agent systems (specialized models for each task) using comprehensive metrics spanning diagnostic outcomes, process adherence, and cost efficiency. Our results reveal a paradox: while multi-agent systems generally outperformed single agents, the component-optimized or \textit{Best of Breed} system with superior components and excellent process metrics (85.5\% information accuracy) significantly underperformed in diagnostic accuracy (67.7\% vs. 77.4\% for a top multi-agent system). This finding underscores that successful integration of AI in healthcare requires not just component level optimization but also attention to information flow and compatibility between agents. Our findings highlight the need for end to end system validation rather than relying on component metrics alone.
\end{abstract}


\maketitle

\section{Introduction}

Artificial intelligence (AI) is rapidly transforming healthcare across diagnosis \cite{zhou2024largelanguagemodelsdisease}, treatment planning \cite{wilhelm2023large}, and patient management \cite{neo2024use}. As AI systems grow in complexity, the focus has shifted from single-model solutions toward networks of specialized models (``agents'') \cite{sapkota2025aiagentsvsagentic} that collaboratively handle different aspects of patient care. Recent studies, including Google DeepMind's AMIE \cite{tu2025towards}, have demonstrated agent-based systems exceeding primary care physicians' performance in randomized clinical settings, while frameworks like MASH \cite{moritz2025coordinated} and CRAFT-MD \cite{johri2024craftmd} have explored both the potential and pitfalls of multi-agent approaches.

Multi-agent AI systems mirror interdisciplinary healthcare teams, where specialists such as radiologists, pathologists, and physicians collaborate to synthesize comprehensive diagnoses. This modular approach can improve interpretability \cite{gordon2024ai}, simplify troubleshooting, and enable task-specific optimization \cite{shu2025unlocking}. However, a critical challenge arises from interactions among individually optimized agents \cite{eliot2024multiagent}. We term this the \textbf{Optimization Paradox:} the phenomenon where excellent performance at the individual agent or component level does not necessarily translate to high overall system performance. This misalignment between individual and system-level effectiveness poses risks to patient safety and clinician trust.

\begin{figure*}[htbp]
  \centering
  \includegraphics[width=0.9 \linewidth]{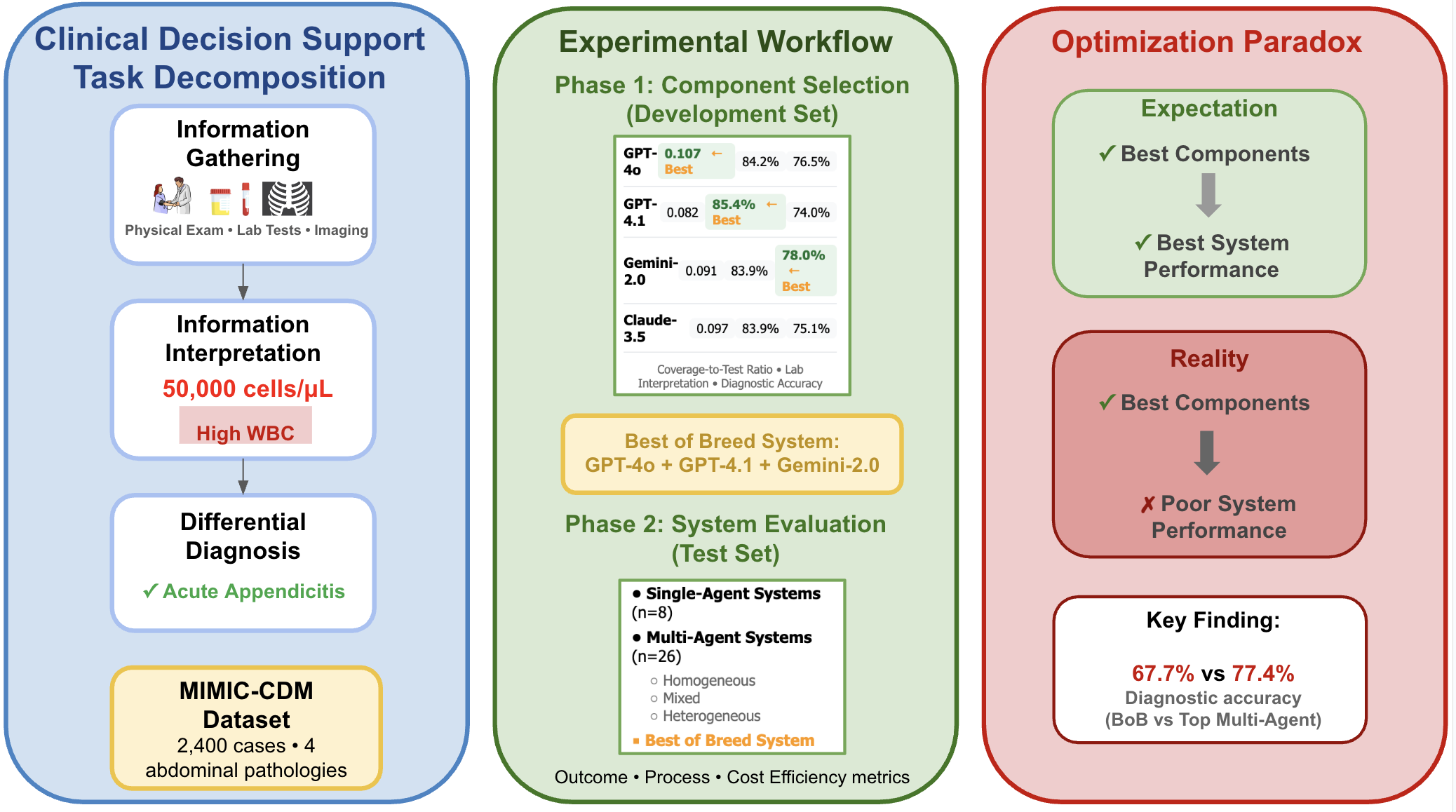} 
  \caption{Overview of experimental methodology and the Optimization Paradox. The Clinical Decision Support Task Decomposition (left) breaks the diagnostic process into three specialized components. The Experimental Workflow (center) shows Phase 1 component selection and Phase 2 system comparison. The Optimization Paradox (right) illustrates the counterintuitive finding: while the \textit{Best of Breed} system was constructed from top-performing components, it achieved poor diagnostic accuracy compared to alternative systems.}
  \label{fig:figure_1} 
\end{figure*}

This study addresses the Optimization Paradox within clinical decision support systems by examining three essential components of the diagnostic process: information gathering (ordering appropriate clinical tests), interpretation (analyzing lab results), and differential diagnosis (identifying potential medical conditions) (Figure \ref{fig:figure_1}). We compare multi-agent systems, where specialized agents manage each task, to single-agent systems, where one model performs all tasks. Our evaluation uses the MIMIC-CDM dataset comprising 2,400 real patient cases across four common abdominal pathologies \cite{hager2024evaluation}. 

We introduce clinically relevant evaluations extending beyond diagnostic accuracy to include process metrics (appropriate test ordering and accurate lab value interpretation) and cost efficiency metrics (clinical resource utilization and computational demands). Our findings reveal that while certain multi-agent systems achieve impressive process-level performance, this does not always translate into clinical effectiveness. The component-optimized or \textit{Best of Breed} system exemplifies this paradox: despite achieving 85.5\% accuracy in lab interpretation, its overall diagnostic accuracy (67.7\%) was significantly lower than a top performing multi-agent system (77.4\%; McNemar's test, p < 0.001) without component optimization. This 10-percentage accuracy drop poses clinically significant risks, potentially increasing misdiagnoses and compromising patient outcomes when AI systems are deployed solely based on component-level validation \cite{Macaluso2012AbdominalPain}.

Our study underscores the necessity of rigorous, end-to-end validation of AI systems prior to clinical implementation, emphasizing that effective patient outcomes depend on careful system-wide integration rather than isolated component excellence. 

\ 

\section{Methods}

This study used the MIMIC-CDM dataset, a curated subset of the MIMIC-IV database containing 2,400 real patient cases from Beth Israel Deaconess Medical Center \cite{hager2024evaluation}. We focused on four common abdominal pathologies: appendicitis (957 cases), pancreatitis (538 cases), cholecystitis (648 cases), and diverticulitis (257 cases). These pathologies were selected based on three criteria: high prevalence at the emergency department (ED) \cite{DiSaverio2020WSES, cervellin2016epidemiology}, high diagnostic complexity due to overlapping clinical presentations \cite{vanBredaVriesman2006Mimics}, and availability of extensive clinical data within MIMIC-CDM for robust case development. This combination creates an ideal testbed for evaluating AI-based differential diagnosis systems. All patients in this cohort presented to the ED with acute abdominal pain and received one of these four conditions as their primary diagnosis.

Each case includes comprehensive clinical data: patient history of present illness, physical examination findings, laboratory results (138,788 lab values from 480 unique tests), imaging reports (5,959 reports including abdominal CT, ultrasound, and X-rays), and procedural information. All available data was de-identified, with any mentions of the primary diagnosis replaced with underscores to prevent trivial pattern matching.

\subsection{Decomposition of the Clinical Decision Support Task}
We adapted MIMIC-CDM's evaluation framework and decomposed the diagnostic workflow into three distinct tasks:
\begin{itemize}
    \item \textbf{Information Gathering:} Requesting relevant clinical information based on patient presentation, including physical examination findings, laboratory tests, and imaging studies.
    
    \item \textbf{Information Interpretation:} Processing and interpreting raw clinical data, such as classifying laboratory results as ``high,'' ``normal,'' or ``low'' relative to reference ranges. 
    
    \item \textbf{Differential Diagnosis:} Synthesizing gathered and interpreted information to generate a ranked list of potential diagnoses through clinical reasoning.
\end{itemize}

We compared two system designs: single-agent systems where one LLM performed all three tasks end-to-end, versus multi-agent systems that divided these tasks among specialized agents.

To maintain data integrity, we employed a \textit{Retriever LLM} that processed information gathering requests and retrieved only specified tests from patient records. GPT-4o was selected for its cost-effectiveness and 100\% retrieval accuracy on the preliminary set. The retrieval prompt is in Appendix \ref{sec:prompt_development}.

\subsection{Data Splits}

 We set aside 20 cases (5 per pathology) as a pilot set for prompt development and pipeline testing during initial data preparation. We excluded these cases from all subsequent model evaluations. Appendix \ref{sec:prompt_development} details the prompts we developed. We then divided the remaining 2,380 cases into two datasets:

\begin{itemize}[topsep=0pt, partopsep=0pt]
   \item \textbf{Development Set:} We used 1,190 cases (50\%), stratified by pathology, for Phase 1 component selection. This identified the best agent for each role to construct the \textit{Best of Breed} system. No training or fine-tuning was performed.
   
   \item \textbf{Test Set:} We used the remaining 1,190 cases (50\%), stratified by pathology, as our held-out test set for Phase 2 evaluation of all systems to assess final performance.
\end{itemize}

\subsection{Model Implementation}

We created agents using LLMs from multiple families including GPT (GPT-4o, GPT-4.1), Claude (Claude-3.5-Sonnet), Gemini (Gemini-1.5-Pro, Gemini-2.0-Flash), Llama (Llama-3.3-70b), and reasoning models (o3-mini, DeepSeek-R1). All experiments were conducted on a PHI-compliant shared cluster. API calls to all models were made through the institution's secure Azure instance, ensuring patient data remained within the institutional environment and maintaining full HIPAA compliance.

\subsection{Evaluation Metrics}
\label{sec:eval_metrics}
We assessed performance using a set of metrics spanning diagnostic outcomes, process adherence, and cost efficiency. These metrics were designed to capture the quality of final diagnostic decisions, the clinical appropriateness of the decision-making process, and resource utilization throughout the workflow.

\subsubsection{\textbf{Outcome Metrics}}
We evaluated diagnostic accuracy across multiple dimensions to assess the quality of clinical reasoning:

\begin{itemize}
    \item \textbf{Overall accuracy:} Micro-averaged (treating each case equally) and macro-averaged (treating each pathology equally) accuracy across all patients.
    
    \item \textbf{Disease-specific accuracy:} Individual accuracy for each of the four target pathologies (appendicitis, pancreatitis, cholecystitis, diverticulitis).
    
    \item \textbf{Top-k accuracy:} Frequency with which the correct diagnosis appeared in the top 1, 3, or 5 positions of the ranked differential diagnosis list, reflecting real-world scenarios where clinicians consider multiple possibilities.
\end{itemize}

\subsubsection{\textbf{Process Metrics}}
We assessed adherence to established clinical guidelines for information gathering and interpretation, which are essential for evidence-based medical practice:

\textbf{Information gathering} was evaluated along four key dimensions:
\begin{itemize}
    \item \textbf{Coverage:} We identified recommendations for physical examination maneuvers (e.g., palpating specific abdominal areas to check for tenderness), laboratory test categories, and imaging modalities for each pathology based on published clinical guidelines (Appendix \ref{sec:rec_tests}) \cite{DiSaverio2020Appendicitis, Pisano2020Cholecystitis, Hall2020Diverticulitis, Leppaniemi2019Pancreatitis, yale2022appendicitis, salati2023cholecystitis, peery2022diverticulosis, tenner2013pancreatitis}. Coverage scores assessed the proportion of recommended categories for which the agent requested at least one item:

$$\text{Coverage Score} = \frac{N_{\text{lab}} + N_{\text{img}} + N_{\text{maneuver}}}{N_{\text{lab\_rec}} + N_{\text{img\_rec}} + N_{\text{maneuver\_rec}}}$$

where $N_{\text{lab}}$, $N_{\text{img}}$, and $N_{\text{maneuver}}$ represent covered laboratory, imaging, and physical examination categories, respectively, and $N_{\text{lab\_rec}}$, $N_{\text{img\_rec}}$, and $N_{\text{maneuver\_rec}}$ represent the total recommended categories for each type. High coverage indicates comprehensive, guideline-concordant information gathering.
    
    \item \textbf{Guideline adherence for physical examination:} Clinical guidelines universally recommend physical examination as the initial diagnostic step for patients with acute abdominal symptoms. We measured the percentage of cases where agents correctly ordered physical examination as their first action.
    
    \item \textbf{Average number of tests per patient:} Average number of diagnostic tests (laboratory, imaging, and physical examination maneuvers) requested per patient, providing insight into resource utilization patterns.
    
    \item \textbf{Coverage-to-test ratio:} Balances comprehensive test assessment with efficient resource utilization:

$$\text{Coverage-to-test ratio} = \frac{\text{Coverage Score}}{\text{Average tests per patient}}$$

This metric rewards systems that achieve high guideline adherence with minimal test ordering, reflecting the clinical imperative to obtain necessary diagnostic information while avoiding unnecessary testing that increases costs and patient burden. To examine this metric's relationship with clinical outcomes, we performed a Spearman correlation analysis between the ratio and final diagnostic accuracy across all systems.
\end{itemize}

\textbf{Information interpretation} was measured as the percentage of lab values correctly classified as ``high,'' ``normal,'' or ``low'' relative to reference ranges. This represents an important clinical skill requiring basic numerical literacy that any competent system should master.

\subsubsection{Cost Efficiency Metrics}
\begin{itemize}
    \item \textbf{Computational cost:} Estimated cost based on token usage and publicly available API pricing for each model.
    
    \item \textbf{Clinical resource cost:} Average Medicare reimbursement cost for all laboratory tests ordered per patient, providing a realistic estimate of healthcare expenditure associated with each system's diagnostic approach. \cite{cms2025clabq2}
\end{itemize}

\subsection{Experiments}

\subsubsection{\textbf{Phase 1: Component Selection and Best of Breed Construction}}  

Individual LLMs were configured as single agents to perform all three tasks end-to-end on the development set (n=1,190). We measured performance across process and outcome metrics (Section~\ref{sec:eval_metrics}) and selected the top-performing agent for each task to construct the \textit{Best-of-Breed (BoB)} system.

\begin{itemize}
   \item \textbf{BoB Information Gathering Agent:} Selected for \textbf{highest coverage-to-test ratio} while maintaining coverage score >0.5. This threshold prevents selection of agents with spuriously high ratios due to low coverage and low average tests per patient.
   
   \item \textbf{BoB Information Interpretation Agent:} Selected for \textbf{highest laboratory interpretation accuracy}, demonstrating superior numerical and contextual analysis of laboratory results.
   
   \item \textbf{BoB Differential Diagnosis Agent:} Selected for \textbf{highest diagnostic accuracy} (micro-averaged/top-1), indicating optimal clinical reasoning for primary diagnosis identification.
\end{itemize}

\begin{figure}[htbp]
  \centering
  \includegraphics[width=0.8\linewidth]{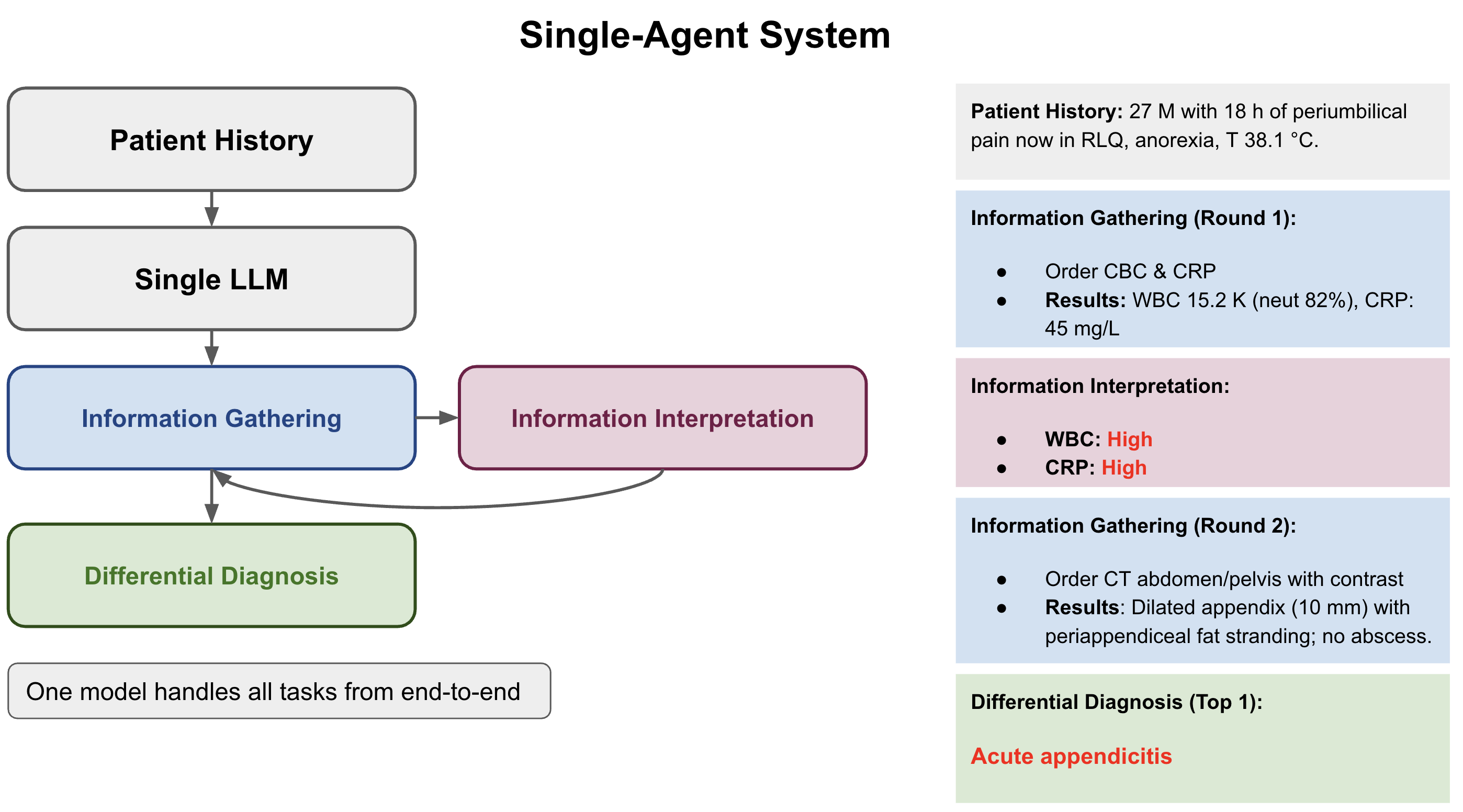} 
  \caption{Single-agent system  (left) and example workflow (right) where one LLM handles the complete clinical workflow from patient history through information gathering, interpretation, and differential diagnosis, demonstrated with an acute appendicitis case.}
  \label{fig:figure_2} 
\end{figure}

\subsubsection{\textbf{Phase 2: System Performance Comparison}}

\begin{figure}[htbp]
  \centering
  \includegraphics[width=0.8\linewidth]{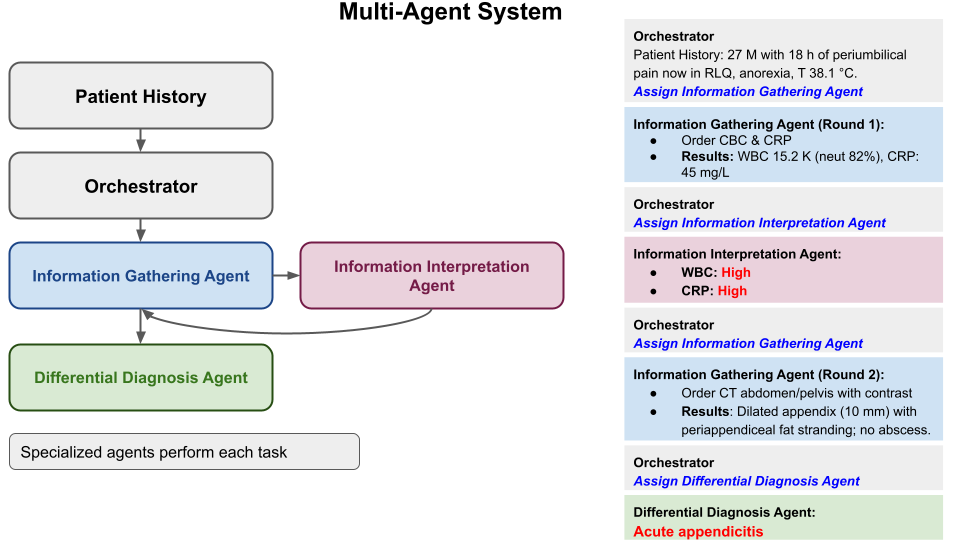} 
  \caption{Multi-agent system (left) and example workflow (right) where an orchestrator coordinates specialized agents for information gathering, interpretation, and differential diagnosis, demonstrated with an acute appendicitis case.}
  \label{fig:figure_3} 
\end{figure}

All systems were evaluated on the held-out test set (n=1,190) to assess real-world performance and investigate the optimization paradox. The evaluation included:

\begin{itemize}
    \item \textbf{Single-Agent Systems:} Individual LLMs performing all three tasks end-to-end, serving as baselines for comparison (Figure \ref{fig:figure_2}).
    
    \item \textbf{Multi-Agent Systems:} Three specialized agents coordinated by a basic orchestrator using conditional logic to route tasks appropriately (Figure \ref{fig:figure_3}). The orchestrator implemented a sequential workflow with explicit handoffs: (1) Information Gathering Agent requests tests, (2) Retriever LLM fetches requested data, (3) Information Interpretation Agent processes lab results, and (4) Differential Diagnosis Agent generates ranked diagnoses. Complete implementation details are provided in Appendix \ref{sec:orchest}. 
    These systems are categorized by their model backbone composition:
    \begin{itemize}
        \item \textbf{Homogeneous:} All three agents use the same backbone (e.g., GPT-4o/GPT-4o/GPT-4o)
        
        \item \textbf{Mixed:} Two agents share a backbone, one differs (e.g., GPT-4o/GPT-4o/Gemini-Flash)
        
        \item \textbf{Heterogeneous:} All three agents use distinct backbones (e.g., GPT-4o/Claude-3.5/Gemini-Flash)
    \end{itemize}
    
    \item \textbf{Best of Breed System:} A heterogeneous multi-agent system constructed using the top-performing agent from Phase 1 for each specialized task.
\end{itemize}

The \textit{Best of Breed} system was created by utilizing the top-performing agents from Phase 1, with each agent assigned to its task of specialization. This resulted in a heterogeneous multi-agent system. Given the computational complexity of evaluating all possible heterogeneous multi-agent permutations, we constructed these systems by selecting the next best performing LLMs for each component, ensuring meaningful diversity.

\

\subsection{Statistical Analysis}

\textbf{Primary Metrics:} We used win rate as our primary performance metric, defined as the percentage of pairwise comparisons where one system type outperforms another. Win rates are robust for small sample sizes and less sensitive to outliers.

\textbf{Statistical Tests:} We applied Mann-Whitney U tests for group comparisons (e.g., single-agent vs. multi-agent systems) and McNemar's test for paired comparisons between specific systems (e.g., \textit{Best of Breed} vs. other systems). We report test statistics, p-values, and 95\% confidence intervals.

\textbf{Effect Size Analysis:} We complemented statistical tests with effect size measures to quantify the magnitude of performance differences:
\begin{itemize}
    \item \textbf{Cohen's d:} For comparing different system types (e.g., single-agent vs. multi-agent)
    
    \item \textbf{Glass's Delta:} Used for comparing a specific agent system (e.g., \textit{Best of Breed}) against a reference group (e.g., all other multi-agent systems), calculated as: 
    $d = \frac{\text{Mean}_{\text{specific}} - \text{Mean}_{\text{reference}}}{\text{SD}_{\text{reference}}}$
\end{itemize}

Effect sizes were interpreted using standard conventions: small ($d = 0.2$), medium ($d = 0.5$), and large ($d = 0.8$) effects. 

\subsection{Code and Data Availability}
All code for implementing the single and multi-agent systems, orchestrator, and evaluation metrics is available at \textcolor{blue}{\url{https://github.com/som-shahlab/opt-paradox}}. The implementation includes prompt templates, orchestration logic, and evaluation scripts to support reproducibility of our findings.

\section{Results}
\subsection{Phase 1: Component Selection and Best of Breed Construction}
We evaluated individual LLMs on all three tasks using the development set (n=1,190). The top performers for each task formed our \textit{Best of Breed} system (Table~\ref{tab:single_agent_performance}).

\begin{itemize}
    \item \textbf{Information Gathering:} GPT-4o achieved the highest coverage-to-test ratio (0.107) with good average coverage per patient (0.64), making it an optimal information gathering agent.
    
    \item \textbf{Information Interpretation:} GPT-4.1 led in laboratory interpretation tasks with 85.4\% accuracy in classifying lab values relative to reference ranges.
    
    \item \textbf{Differential Diagnosis:} Gemini-2.0-Flash demonstrated superior micro-averaged diagnostic accuracy (78\%).
\end{itemize}

Based on these evaluations, we constructed the \textit{Best of Breed} system using GPT-4o for information gathering, GPT-4.1 for information interpretation, and Gemini-2.0-Flash for differential diagnosis.

\

\begin{table}[htbp]
\centering
\small
\begin{tabular}{lccc}
\hline
\textbf{LLM} & \textbf{Info Gathering} & \textbf{Info Interpretation} & \textbf{Diagnosis} \\
\hline
Gemini-2.0-Flash & 0.091 & 83.9 & \textbf{77.98} \\
GPT-4o & \textbf{0.107} & 84.2 & 76.47 \\
Claude-3.5-Sonnet & 0.097 & 83.9 & 75.13 \\
Llama-3.3-70b & 0.088 & 75.9 & 74.96 \\
DeepSeek-R1 & 0.086 & 71.9 & 74.87 \\
GPT-4.1 & 0.082 & \textbf{85.4} & 73.95 \\
Gemini-1.5-Pro & 0.082 & 79.8 & 72.77 \\
o3-mini & 0.081 & 81.6 & 61.6 \\
\hline
\end{tabular}
\caption{Task-level performance for single-agents with the best value for each task highlighted in \textbf{bold}. Coverage-to-test ratio measures information gathering, lab interpretation accuracy reflects information interpretation, and micro-averaged diagnostic accuracy represents differential diagnosis capability.}
\label{tab:single_agent_performance}
\end{table}

\subsection{Phase 2: System Performance Comparison}

A total of 8 single-agent and 26 multi-agent systems were evaluated on the held-out test set (Table~\ref{tab:performance_comparison}).

\begin{table*}[htbp]
\centering
\begin{tabular}{lccc}
\toprule
\textbf{Metric Category} & 
\begin{tabular}{@{}c@{}}
\textbf{Multi-Agent Systems vs.} \\
\textbf{Single-Agent Systems} \\
\textbf{(Win Rate for Multi-Agents)}
\end{tabular} & 
\begin{tabular}{@{}c@{}}
\textbf{BoB System vs. Other} \\
\textbf{Multi-Agent Systems} \\
\textbf{(Win Rate for BoB)}
\end{tabular} & 
\begin{tabular}{@{}c@{}}
\textbf{BoB System vs.} \\
\textbf{Single-Agent Systems} \\
\textbf{(Win Rate for BoB)}
\end{tabular} \\
\midrule
Information Gathering & 
\textbf{78.4\%} (Cohen's $d$ = 1.06) & 
\textbf{80.0\%} (Glass's $\Delta$ = 1.29) & 
\textbf{100\%} (Glass's $\Delta$ = 1.66) \\
Information Interpretation & 
\textbf{87.5\%} (Cohen's $d$ = 1.79) & 
\textbf{76.0\%} (Glass's $\Delta$ = 0.60) & 
\textbf{100\%} (Glass's $\Delta$ = 1.44) \\
Diagnosis Accuracy & 
52.9\% (Cohen's $d$ = 0.05) & 
\textbf{8.0\%} (Glass's $\Delta$ = $-$1.04) & 
\textbf{12.5\%} (Glass's $\Delta$ = $-$1.15) \\
Computational Cost & 
\textbf{76.9\%} (Cohen's $d$ = 0.96) & 
\textbf{80.0\%} (Glass's $\Delta$ = 0.81) & 
\textbf{87.5\%} (Glass's $\Delta$ = 0.92) \\
Clinical Resource Cost & 
60.1\% (Cohen's $d$ = 0.31) & 
\textbf{84.0\%} (Glass's $\Delta$ = 1.46) & 
\textbf{87.5\%} (Glass's $\Delta$ = 1.31) \\
\bottomrule
\end{tabular}
\caption{Win rates (\%) between multi-agent, BoB, and single-agent systems across multiple evaluation metrics. Coverage-to-test ratio measures information gathering, lab interpretation accuracy reflects information interpretation , and micro-averaged diagnostic accuracy represents differential diagnosis capability. Effect sizes are reported using Cohen's $d$ for parametric comparisons and Glass's $\Delta$ for non-parametric comparisons (\textbf{Bold indicates large or medium effect size, $|$Cohen's $d|$ $\geq 0.5$ or $|$Glass's $\Delta|$ $\geq 0.5$}).}
\label{tab:performance_comparison}
\end{table*}

\subsubsection{\textbf{Multi-Agent vs Single-Agent Systems}}

Multi-agent systems (n=26) significantly outperformed single-agent systems (n=8) on process metrics, winning 78.4\% of pairwise comparisons for information gathering (p = 0.015), 87.5\% for information interpretation (p = 0.0008), and 76.9\% for computational cost (p = 0.022). However, multi-agent system advantages were modest and non-significant for diagnostic accuracy (52.9\% win rate) and clinical resource costs (60.1\% win rate). Thus, multi-agent systems enhance process quality and computational efficiency but show limited improvement in clinical outcomes. Notably, the coverage-to-test ratio showed no significant correlation with diagnostic accuracy (Spearman's $\rho = -0.057$, $p = 0.748$), underscoring the disconnect between process metrics and outcomes.
All raw numbers can be found in Appendix \ref{sec:test}

\subsubsection{\textbf{Best of Breed vs Multi-Agent Systems: The Optimization Paradox
}}

The Optimization Paradox emerged when comparing the \textit{Best of Breed} system against other multi-agent systems (n=25). While BoB achieved high win rates across process metrics including information gathering (80.0\%), information interpretation (76.0\%), computational cost (80.0\%), and clinical resource cost (84.0\%), it underperformed in diagnostic accuracy with only an 8.0\% win rate. 

Direct comparison with the top-performing multi-agent system revealed that BoB's diagnostic accuracy (67.65\%) was significantly lower than the baseline (77.39\%), representing a 9.75\% decrease (95\% CI: 7.11\% to 12.38\%; McNemar's test, p < 0.0001). This contrast between strong component performance and poor diagnostic outcomes demonstrates the Optimization Paradox: optimizing individual components can undermine overall system performance.

\subsubsection{\textbf{Best of Breed vs Single-Agent Systems}}

The Optimization Paradox became even more pronounced when comparing the \textit{Best of Breed} system against single-agent systems (n=8). While BoB achieved perfect win rates across process metrics including information gathering (100\%) and information interpretation (100\%), along with strong performance in cost efficiency metrics including computational cost (87.5\% win rate) and clinical resource cost (87.5\% win rate), it critically underperformed in diagnostic accuracy with only a 12.5\% win rate.

Direct comparison with the top-performing single-agent system revealed that BoB's diagnostic accuracy (67.65\%) was significantly lower than the baseline (75.63\%), representing a 7.98\% decrease (95\% CI: 5.39\% to 10.57\%; McNemar's test, p < 0.0001). This contrast between strong operational metrics and poor diagnostic outcomes further confirms that optimizing individual components can undermine overall system performance.

\subsubsection{\textbf{Model Backbone Effects on Diagnostic Performance
}}

The Optimization Paradox in our \textit{Best of Breed} system prompted investigation into whether diagnostic performance relates to model diversity within multi-agent systems. We compared diagnostic accuracy across homogeneous systems (n=7, all agents from same backbone), mixed systems (n=12, two agents from one backbone), and heterogeneous systems (n=6, all agents from different backbones, including BoB). We excluded one extreme outlier (DeepSeek system with 54\% accuracy) from the homogeneous group analysis.

Homogeneous systems (median = 74.29\%) showed no significant difference from mixed systems (median = 74.75\%; p = 0.967, Cohen's d = 0.019). However, heterogeneous systems (median = 71.22\%) demonstrated lower performance than both homogeneous (p = 0.073, Cohen's d = 1.41) and mixed systems (p = 0.039, Cohen's d = 1.17). While p-values did not reach Bonferroni-corrected significance ($\alpha = 0.0167$), the large effect sizes suggest heterogeneous compositions face inherent diagnostic challenges, potentially explaining the Optimization Paradox.

\subsubsection{\textbf{Why Best of Breed Fails: Information Flow Breakdown}}

To understand why the \textit{Best of Breed} system failed despite superior component metrics, we conducted systematic error analysis on all 1,190 test cases, comparing failure patterns against the top-performing multi-agent system (Table~\ref{tab:failure_analysis}). 

\ 

\begin{table}[htbp]
\centering
\small
\begin{tabular}{lcc}
\toprule
\textbf{Metric} & \textbf{BoB} & \textbf{Top Multi-Agent} \\
\midrule
\multicolumn{3}{l}{\textit{Overall Performance}} \\
Hallucinated Test Results & 165 (13.87\%) & 5 (0.42\%) \\
Unauthorized Test Ordering & 165 (13.87\%) & 9 (0.76\%) \\
Insufficient Info Gathering & 84 (7.06\%) & 23 (1.93\%) \\
\midrule
\multicolumn{3}{l}{\textit{Head-to-Head Comparison}} \\
Hallucinated Test Results & 81 (46.55\%) & 0 (0.00\%) \\
Insufficient Info Gathering & 63 (36.21\%) & 0 (0.00\%) \\
Other Failures & 30 (17.24\%) &   0 (0.00\%) \\
\bottomrule
\end{tabular}
\caption{System failure comparison showing overall performance (1,190 cases) and head-to-head analysis of 174 cases where the top-performing multi-agent system succeeded but Best of Breed failed.}
\label{tab:failure_analysis}
\end{table}

\paragraph{Information Gathering Failures:}
BoB's information gathering agent (GPT-4o) exhibited critical failure patterns, including insufficient information gathering in 7.06\% of cases where the agent concluded test gathering despite missing essential diagnostic information, particularly imaging tests in pancreatitis cases.

\paragraph{Diagnosis Agent Failures:}
These information deficits triggered compensatory behaviors in BoB's diagnosis agent (Gemini-2.0-Flash). When faced with insufficient data, the agent violated protocol by attempting unauthorized test ordering in 13.87\% of cases, and then hallucinated test results at the same rate. This represents a serious safety failure in clinical decision-making.

\paragraph{Top-Performing Multi-Agent System Success:}
In contrast, the top-performing multi-agent system (using Gemini-2.0-Flash for information gathering and interpretation and GPT-4o for diagnosis) demonstrated superior information flow management with lower failure rates: only 1.93\% insufficient information gathering, 0.76\% unauthorized test ordering, and 0.42\% hallucinated results, representing a \textbf{33-fold} reduction in hallucination compared to BoB.

\paragraph{Head-to-Head Analysis:}
Direct comparison of 174 cases where the top-performing multi-agent system succeeded but BoB failed revealed that nearly half (46.55\%) involved test result hallucination, while 36.21\% showed insufficient information gathering. These findings demonstrate that the Optimization Paradox stems from fundamental agent compatibility issues rather than individual component deficiencies.

These results show that component-level metrics cannot capture agent interactions. Individually superior agents may create systematic coordination failures when combined, undermining overall performance despite strong standalone capabilities.

\

\section{Discussion}

Our study reveals a striking \textbf{Optimization Paradox} in multi-agent systems: the \textit{Best of Breed} system, built from top-performing components, excelled in process and cost efficiency metrics yet achieved only 67.7\% diagnostic accuracy. This level of performance, coupled with test result hallucination in 13.87\% of cases, is clinically unacceptable and represents a serious safety hazard, potentially leading to delayed or incorrect treatment, unnecessary procedures, and adverse outcomes \cite{Macaluso2012AbdominalPain}. This paradox demonstrates that successful multi-agent systems require not just component optimization but careful attention to information flow between agents.

Several factors explain this surprising outcome. First, our process metrics captured whether agents followed guidelines but missed diagnostic relevance. Thus, the \textit{Best of Breed} system efficiently followed clinical guidelines, but these metrics failed to assess whether collected data matched the diagnostic agent's specific requirements. 

Second, the diagnostic agent showed poor adaptability when processing information from unfamiliar upstream partners. During Phase 1 evaluation, it performed well on its specialized task. When combined with other top-performing agents in the \textit{Best of Breed} system in Phase 2, the information flow and formatting patterns were disrupted, causing systematic diagnostic failures. This highlights that multi-agent systems require holistic evaluation rather than component-level optimization.

Our backbone composition analysis reveals the underlying mechanism: while mixed systems (two identical + one different model) performed as well as homogeneous systems, heterogeneous systems like \textit{Best of Breed} showed significant degradation. The coordination challenges likely stem from fundamental differences in how model backbones process and communicate information. Each backbone (GPT, Claude, Gemini) has distinct training approaches, prompt sensitivity patterns, and output formatting preferences. When these different ``communication styles'' interact, information may be lost or misinterpreted during handoffs. Our error analysis confirms this mechanism: in the 174 cases where the top-performing system succeeded but \textit{Best of Breed} failed, nearly half (46.55\%) involved dangerous test result hallucination, while the compatible agents showed no hallucination failures.

These technical findings have important practical implications for AI deployment. Healthcare organizations should exercise caution when adopting modular AI solutions, as component metrics poorly predict integrated performance. Procuring best-in-class point solutions for each task while expecting seamless integration can create nominally efficient but ultimately ineffective and potentially unsafe pipelines. End-to-end validation against clinical outcomes is essential before deployment. In addition, regulatory frameworks should require system-level performance evidence rather than relying solely on component accuracy metrics.

Several limitations warrant consideration. The dataset (2,400 cases from a single academic center) limits generalizability across institutions and patient populations, and lacks external validation. Our focus on four abdominal pathologies may not generalize to other clinical domains. Additionally, our component selection prioritized single metrics per task rather than multi-dimensional optimization strategies, and our multi-agent system used basic orchestration without iterative reasoning or dynamic agent communication.

Future work should develop process metrics that better correlate with clinical outcomes, investigate selection methods that optimize for system-level performance rather than isolated component excellence, and explore dynamic agent architectures capable of iterative reasoning and self-correction. Most importantly, external validation across diverse clinical settings is needed to establish the generalizability of the Optimization Paradox.
\bibliographystyle{unsrt}
\bibliography{software}

\begin{thebibliography}{10}

\bibitem{zhou2024largelanguagemodelsdisease}
Shuang Zhou, Zidu Xu, Mian Zhang, Chunpu Xu, Yawen Guo, Zaifu Zhan, Sirui Ding, Jiashuo Wang, Kaishuai Xu, Yi~Fang, Liqiao Xia, Jeremy Yeung, Daochen Zha, Genevieve~B. Melton, Mingquan Lin, and Rui Zhang.
\newblock Large language models for disease diagnosis: A scoping review, 2024.

\bibitem{wilhelm2023large}
Theresa~Isabelle Wilhelm, Jonas Roos, and Robert Kaczmarczyk.
\newblock Large language models for therapy recommendations across 3 clinical specialties: Comparative study.
\newblock {\em Journal of Medical Internet Research}, 25:e49324, Oct 2023.

\bibitem{neo2024use}
Jin Rui~Edmund Neo, Joon~Sin Ser, and San~San Tay.
\newblock Use of large language model-based chatbots in managing the rehabilitation concerns and education needs of outpatient stroke survivors and caregivers.
\newblock {\em Frontiers in Digital Health}, 6:1395501, May 2024.

\bibitem{sapkota2025aiagentsvsagentic}
Ranjan Sapkota, Konstantinos~I. Roumeliotis, and Manoj Karkee.
\newblock Ai agents vs. agentic ai: A conceptual taxonomy, applications and challenges, 2025.

\bibitem{tu2025towards}
Tao Tu, Mike Schaekermann, Anil Palepu, Khaled Saab, Jan Freyberg, Ryutaro Tanno, Amy Wang, Brenna Li, Mohamed Amin, Yong Cheng, Elahe Vedadi, Nenad Tomasev, Shekoofeh Azizi, Karan Singhal, Le~Hou, Albert Webson, Kavita Kulkarni, S.~Sara Mahdavi, Christopher Semturs, Juraj Gottweis, Joelle Barral, Katherine Chou, Greg~S. Corrado, Yossi Matias, and Vivek Natarajan.
\newblock Towards conversational diagnostic artificial intelligence.
\newblock {\em Nature}, April 2025.
\newblock Published online 09 April 2025.

\bibitem{moritz2025coordinated}
Michael Moritz, Eric Topol, and Pranav Rajpurkar.
\newblock Coordinated ai agents for advancing healthcare.
\newblock {\em Nature Biomedical Engineering}, 2025.
\newblock Commentary.

\bibitem{johri2024craftmd}
Shreya Johri, Jaehwan Jeong, Benjamin~A. Tran, Daniel~I Schlessinger, Shannon Wongvibulsin, Zhuo~Ran Cai, Roxana Daneshjou, and Pranav Rajpurkar.
\newblock {CRAFT}-{MD}: A conversational evaluation framework for comprehensive assessment of clinical {LLM}s.
\newblock In {\em AAAI 2024 Spring Symposium on Clinical Foundation Models}, 2024.

\bibitem{gordon2024ai}
Rachel Gordon.
\newblock Ai agents help explain other ai systems, January 2024.
\newblock MIT News, Massachusetts Institute of Technology.

\bibitem{shu2025unlocking}
Raphael Shu, Yi~Zhang, Michelle Yuan, Nilaksh Das, and Monica Sunkara.
\newblock Unlocking complex problem-solving with multi-agent collaboration on amazon bedrock.
\newblock \url{https://aws.amazon.com/blogs/machine-learning/unlocking-complex-problem-solving-with-multi-agent-collaboration-on-amazon-bedrock/}, January 2025.
\newblock AWS Machine Learning Blog.

\bibitem{eliot2024multiagent}
Lance~B. Eliot.
\newblock Multi-agent ai orchestration shaping up but here’s why it might not be fully shipshape.
\newblock {\em Forbes}, November 2024.
\newblock Innovation, AI.

\bibitem{hager2024evaluation}
Paul Hager, Friederike Jungmann, Robbie Holland, Kunal Bhagat, Inga Hubrecht, Manuel Knauer, Jakob Vielhauer, Marcus Makowski, Rickmer Braren, Georgios Kaissis, and Daniel Rueckert.
\newblock Evaluation and mitigation of the limitations of large language models in clinical decision-making.
\newblock {\em Nature Medicine}, 30:2613--2622, July 2024.

\bibitem{Macaluso2012AbdominalPain}
Christopher~R. Macaluso and Robert~M. McNamara.
\newblock Evaluation and management of acute abdominal pain in the emergency department.
\newblock {\em International Journal of General Medicine}, 5:789--797, 2012.

\bibitem{DiSaverio2020WSES}
Salomone Di~Saverio, Mauro Podda, Belinda De~Simone, Marco Ceresoli, Goran Augustin, Antonio Gori, Luca Ansaloni, Marja Boermeester, Massimo Sartelli, Federico Coccolini, and Fausto Catena.
\newblock Diagnosis and treatment of acute appendicitis: 2020 update of the wses jerusalem guidelines.
\newblock {\em World Journal of Emergency Surgery}, 15(1):27, 2020.

\bibitem{cervellin2016epidemiology}
Gianfranco Cervellin, Riccardo Mora, Andrea Ticinesi, Tiziana Meschi, Ivan Comelli, Fausto Catena, and Giuseppe Lippi.
\newblock Epidemiology and outcomes of acute abdominal pain in a large urban emergency department: retrospective analysis of 5,340 cases.
\newblock {\em Emergency and Critical Care Medicine}, 4(19):1--8, 2016.
\newblock Available online October 15, 2016.

\bibitem{vanBredaVriesman2006Mimics}
Adriaan~C. van Breda~Vriesman and Julien B. C.~M. Puylaert.
\newblock Mimics of appendicitis: Alternative nonsurgical diagnoses with sonography and ct.
\newblock {\em AJR American Journal of Roentgenology}, 186(4):1103--1112, 2006.

\bibitem{DiSaverio2020Appendicitis}
Salomone Di~Saverio, Mauro Podda, Belinda De~Simone, Marco Ceresoli, Goran Augustin, Antonio Gori, Luca Ansaloni, Marja Boermeester, Massimo Sartelli, Federico Coccolini, and Fausto Catena.
\newblock Diagnosis and treatment of acute appendicitis: 2020 update of the wses jerusalem guidelines.
\newblock {\em World Journal of Emergency Surgery}, 15:27, 2020.

\bibitem{Pisano2020Cholecystitis}
Michele Pisano, Nadia Allievi, Kurinchi Gurusamy, Giovanni Borzellino, Carlos~A. Gomes, Andrew~W. Kirkpatrick, and et~al.
\newblock 2020 world society of emergency surgery updated guidelines for the diagnosis and treatment of acute calculus cholecystitis.
\newblock {\em World Journal of Emergency Surgery}, 15:61, 2020.

\bibitem{Hall2020Diverticulitis}
John~F. Hall, Patricia~L. Roberts, Ramon Ricciardi, Thomas~E. Read, Benjamin~R. Davis, Paul~W. Marcello, and Scott~R. Steele.
\newblock The american society of colon and rectal surgeons clinical practice guidelines for the treatment of left-sided colonic diverticulitis.
\newblock {\em Diseases of the Colon \& Rectum}, 63(5):728--747, 2020.

\bibitem{Leppaniemi2019Pancreatitis}
Ari Leppäniemi, Matti Tolonen, Antonio Tarasconi, Gabriele Anania, Helena Segovia-Lohse, Edoardo Gamberini, and et~al.
\newblock 2019 wses guidelines for the management of severe acute pancreatitis.
\newblock {\em World Journal of Emergency Surgery}, 14:27, 2019.

\bibitem{yale2022appendicitis}
Steven~H Yale, Halil Tekiner, and Elizabeth~S Yale.
\newblock Signs and syndromes in acute appendicitis: A pathophysiologic approach.
\newblock {\em World Journal of Gastrointestinal Surgery}, 14(7):727--730, 2022.

\bibitem{salati2023cholecystitis}
Sajad~Ahmad Salati, Khalid Alkhalifah, and Abdul Majeed~Salem Almousa.
\newblock Eponymous signs of acute cholecystitis – a review.
\newblock {\em Saudi Medical Journal Students}, 3(2):44--55, 2023.

\bibitem{peery2022diverticulosis}
Anne~F Peery, Temitope~O Keku, Joseph~A Galanko, and Robert~S Sandler.
\newblock Colonic diverticulosis is not associated with painful abdominal symptoms in a us population.
\newblock {\em Gastro Hep Advances}, 1(1):15--22, 2022.

\bibitem{tenner2013pancreatitis}
Scott Tenner, John Baillie, John DeWitt, and Santhi~Swaroop Vege.
\newblock American college of gastroenterology guideline: management of acute pancreatitis.
\newblock {\em The American Journal of Gastroenterology}, 108(9):1400--1415, 2013.

\bibitem{cms2025clabq2}
{Centers for Medicare \& Medicaid Services}.
\newblock {CY 2025 Q2 Release: Added for April 2025. The update includes all changes identified in CR 13966}.
\newblock \url{https://www.cms.gov/medicare/payment/clinical-laboratory-fee-schedule/clfs-quarterly-update-files}, April 2025.
\newblock File name: 25CLABQ2. Contains 2,105 records.

\end{thebibliography}

\section{Appendix}

\subsection{Prompt Development}
\label{sec:prompt_development}

\begin{figure}[H]            
\centering
\begin{tcolorbox}[
    colback = lightgray,
    colframe = darkgray,
    width   = \columnwidth,
    boxrule = 0.5pt,
    arc     = 3pt,
    left=3pt,right=3pt,top=6pt,bottom=6pt]
\textbf{Single-agent:}

\begin{Verbatim}[fontsize=\scriptsize]
# SYSTEM PROMPT
"""
You are a medical‑AI assistant helping a physician diagnose and treat patients.  **Always follow the exact output formats below.**
------------------------------------------------------------------------
FORMAT 1  (when you still need more information)
Thought: <your reasoning about what information is needed and why>

[If the immediately‑preceding message is a Tool output that contains laboratory values, INSERT the next section exactly once.]

Lab Interpretation: {
    "test_name": {"value": <number>, "interpretation": "high/normal/low"},
    ...
}
Action: <one of: Physical Examination | 
Laboratory Tests | Imaging>
Action Input: <comma‑separated list of specific tests, imaging studies or physical exam maneuver you are requesting.>
IMPORTANT: You can only request one action type at a time. Do not combine multiple action types.
------------------------------------------------------------------------
FORMAT 2  (when you are ready to give the final answer)
Thought: <your complete clinical reasoning>
**Final Diagnosis (ranked):** 
1. <most likely diagnosis>
2. <second most likely diagnosis>
3. <third most likely diagnosis>
4. <fourth most likely diagnosis>
5. <fifth most likely diagnosis>
Treatment: <detailed evidence‑based treatment plan>

**IMPORTANT: After providing FORMAT 2, your task is COMPLETE. Do NOT request any further actions or tools.  FORMAT 2 is the FINAL output. 
Once you provide FORMAT 2, the conversation ENDS.**
------------------------------------------------------------------------
HARD RULES  (read carefully)

1. **Mandatory Lab Interpretation**  
   • If the last message you received is a Tool output with lab data, you MUST include the “Lab Interpretation” JSON block.  
   • If you omit it, your answer will be rejected and you will be asked to try again.

2. JSON validity  
   • The Lab Interpretation block must be valid JSON (double quotes, no trailing commas).  
   • Include both the numeric value and the interpretation (“high”, “normal”, or “low”) for every test you mention.

3. Do NOT mix elements from different formats.

4. “Action Input” is **only** for naming new tests or imaging studies you want to order.  Never place results or interpretations there.

5. **Action Input Content:** The "Action Input" field should ONLY contain a comma-separated list  of test names, imaging studies, or 
physical exam maneuvers. Do NOT include any thoughts, reasoning, interpretations, or other text in the "Action Input" field.

6. **STOP AFTER FORMAT 2:** Once you have provided FORMAT 2 
(Final Diagnosis and Treatment), you MUST stop. Do NOT ask for any more information or tools after FORMAT 2. 

7. Stop asking for additional information 
when you are confident enough to provide FORMAT 2.
------------------------------------------------------------------------
EXAMPLES

Lab Interpretation: {
    "WBC":  {"value": 12.5, "interpretation": "high"},
    "CRP":  {"value": 5.0,  "interpretation": "normal"}
}
Action: Laboratory Tests
Action Input: Serum Lipase, Abdominal Ultrasound
Action: Physical Examination
Action Input: McBurney's Point Tenderness
"""

\end{Verbatim}
\end{tcolorbox}
\caption{Prompts used for the single-agent systems.}
\end{figure}

\begin{figure}[H]            
\centering
\begin{tcolorbox}[
    colback = lightgray,
    colframe = darkgray,
    width   = \columnwidth,
    boxrule = 0.5pt,
    arc     = 3pt,
    left=6pt,right=6pt,top=6pt,bottom=6pt]
\textbf{Multi-agent:}

\begin{Verbatim}[fontsize=\scriptsize]
# INFORMATION GATHERING AGENT

INFO_GATHERING_PROMPT = """\
You are a medical-AI assistant helping a physician COLLECT information that will later be used to diagnose and treat the patient.  
**Always follow the exact output formats below.**
------------------------------------------------------------------------
FORMAT 1  (when you still need more information)

Thought: <your reasoning about what information is needed and why>
Action: <one of: Physical Examination | Laboratory Tests | Imaging>
Action Input: <comma-separated list of specific tests, imaging studies or physical exam maneuver you are requesting.>

IMPORTANT: You can only request one action type at a time. Do not combine multiple action types.
------------------------------------------------------------------------
FORMAT 2  (when you are done collecting information)

Thought: <your complete clinical reasoning>
Action: done
Action Input: ""

**IMPORTANT: After providing FORMAT 2, your task is COMPLETE. Do NOT request any further actions or tools.  
FORMAT 2 is the FINAL output. Once you provide FORMAT 2, the conversation ENDS.**

------------------------------------------------------------------------
HARD RULES  (read carefully)

1. Do NOT mix elements from different formats.

2. “Action Input” is **only** for naming new tests or imaging studies you want to order.  Never place results or interpretations there.

3. **Action Input Content:** The "Action Input" field should ONLY contain a comma-separated list of test names, imaging studies, 
or physical exam maneuvers. 
Do NOT include any thoughts, reasoning, interpretations, or other text in the "Action Input" field.

4. **STOP AFTER FORMAT 2:** Once you have provided FORMAT 2, you MUST stop. Do NOT ask for any more information or tools after FORMAT 2. 

5. Stop asking for additional information when you are confident enough to provide FORMAT 2.
"""

# INFORMATION INTERPRETATION AGENT
INTERPRETATION_PROMPT = """\
You are a medical-AI assistant helping a physician interpret laboratory results that have already been retrieved.  
**Always follow the exact output formats below.**

------------------------------------------------------------------------
FORMAT  (interpret the lab panel you just received)

[If the immediately-preceding message is a Tool output that contains laboratory values, INSERT the next section exactly once.]

Lab Interpretation: {
    "test_name": {"value": <number>, "interpretation": "high/normal/low"},
    ...
}

**IMPORTANT: After providing this FORMAT, your task is COMPLETE. 
Do NOT request any further actions or tools.  This FORMAT is the FINAL output. Once you provide this FORMAT, the conversation ENDS.**
HARD RULES  (read carefully)

1. **Mandatory Lab Interpretation**  
   • If the last message you received is a Tool output with lab data, you MUST include the “Lab Interpretation” JSON block 
   • If you omit it, your answer will be rejected and you will be asked
     to try again.

2. JSON validity  
   • The Lab Interpretation block must be valid JSON (double quotes,
     no trailing commas).  
   • Include both the numeric value and the interpretation
     (“high”, “normal”, or “low”) for every test you mention.

3. Do NOT mix elements from different formats.
"""
\end{Verbatim}
\end{tcolorbox}
\caption{Prompts used for the multi-agent systems.}
\end{figure}

\begin{figure}[H]            
\centering
\begin{tcolorbox}[
    colback = lightgray,
    colframe = darkgray,
    width   = \columnwidth,
    boxrule = 0.5pt,
    arc     = 3pt,
    left=6pt,right=6pt,top=6pt,bottom=6pt]
\textbf{Retriever LLM:}

\begin{Verbatim}[fontsize=\scriptsize]

LABS_MATCHER_PROMPT = """
Available laboratory tests and their results: {available_tests}.
Requested tests: {requested_tests}.

Please retrieve and return the results for the requested tests.
Return each test name along with its corresponding result.
If a test is not available, state that.
Respond in natural language
"""

IMAGING_MATCHER_PROMPT = """
Available imaging studies: {available_imaging}.
Requested imaging: {requested_imaging}.

Please retrieve and return the full report only 
for the imaging study that best matches the 
requested imaging from the available list. 
If the requested imaging is not available, state that. 
Do not propose or mention any additional or alternative tests or imaging. 
Return the study name along with the full report. 
Respond in natural language.
"""

\end{Verbatim}
\end{tcolorbox}
\caption{Prompts used for the retriever LLM.}
\end{figure}

\subsection{Guideline recommended tests}
\label{sec:rec_tests}

\begin{table}[H]
\centering
\small
\begin{tabular}{p{1.8cm}p{2.5cm}p{3.2cm}}
\toprule
\textbf{Pathology} & \textbf{Physical Exam Maneuver} & \textbf{Synonyms} \\
\midrule
Appendicitis & McBurney's Point Tenderness & 
mcburney, mcburney's, mcburney point, mcburney's point, point of mcburney, mcburney tenderness, right iliac tenderness, tenderness at mcburney, tenderness at mcburney's point \\
\midrule
Cholecystitis & Murphy's Sign & 
murphy, murphy's, murphy sign, murphy's sign, inspiratory arrest, halted inspiration, interruption of breath, breath catching, respiratory arrest with palpation \\
\midrule
Diverticulitis & Left Lower Quadrant Tenderness & 
left lower quadrant, llq, sigmoid, sigmoid tenderness, tenderness over sigmoid, left iliac fossa, lif, left-sided abdominal tenderness, sigmoid colon tenderness \\
\midrule
Pancreatitis & Epigastric Tenderness & 
epigastric, epigastrium, upper abdominal, mid-upper abdomen, central upper abdomen, transabdominal tenderness, midline upper abdomen, central abdominal tenderness, mid-epigastric \\
\bottomrule
\end{tabular}
\caption{Physical Examination Maneuvers and Synonyms by Pathology}
\label{tab:physical_exam}
\end{table}

\begin{table}[H]
\centering
\small
\begin{tabular}{p{2.5cm}p{5cm}}
\toprule
\textbf{Pathology} & \textbf{Recommended Tests} \\
\midrule
Appendicitis & 
\textbf{Inflammation:}
\begin{itemize}[leftmargin=*, nosep, itemsep=0pt]
\item White Blood Cell Count (WBC)
\item C-Reactive Protein (CRP)
\end{itemize} \\
\midrule
Cholecystitis & 
\textbf{Inflammation:}
\begin{itemize}[leftmargin=*, nosep, itemsep=0pt]
\item White Blood Cell Count (WBC)
\item C-Reactive Protein (CRP)
\end{itemize}
\textbf{Assess the risk of Chronic Bile Duct Stones (CBDS):}
\begin{itemize}[leftmargin=*, nosep, itemsep=0pt]
\item Alanine Transaminase (ALT)
\item Aspartate Transaminase (AST)
\item Alkaline Phosphatase (ALP)
\item Gamma Glutamyltransferase (GGT)
\item Bilirubin
\end{itemize} \\
\midrule
Diverticulitis & 
\textbf{Inflammation:}
\begin{itemize}[leftmargin=*, nosep, itemsep=0pt]
\item White Blood Cell Count (WBC)
\item C-Reactive Protein (CRP) (predicts severity)
\end{itemize} \\
\midrule
Pancreatitis & 
\textbf{Serum pancreatic enzyme:}
\begin{itemize}[leftmargin=*, nosep, itemsep=0pt]
\item Lipase
\item Amylase
\end{itemize}
\textbf{Other:}
\begin{itemize}[leftmargin=*, nosep, itemsep=0pt]
\item C-Reactive Protein (CRP)
\item Hematocrit
\item Blood Urea Nitrogen (BUN)
\item Procalcitonin
\item serum triglyceride and calcium levels (in absence of gallstones or significant alcohol use)
\end{itemize} \\
\bottomrule
\end{tabular}
\caption{Recommended Lab Tests by Pathology}
\label{tab:lab_tests}
\end{table}

\subsection{Orchestrator Implementation}
\label{sec:orchest}

The orchestrator coordinated specialized agents through a sequential workflow built on LangGraph. Each agent received the patient's clinical context and complete conversation history, enabling informed decision-making at each step. The Information Gathering Agent iteratively requested clinical tests until signaling completion with "Action: done" or reaching the 10-turn limit, at which point control passed to subsequent agents.
Data retrieval was handled through the RetrieveResults tool, which processed three types of requests: physical examinations, laboratory tests, and imaging studies. When agents requested specific tests using natural language (e.g., "Complete blood count, C-reactive protein"), GPT-4o served as a retriever to identify and return the relevant patient data from clinical records.

\begin{figure}[H]            
\centering
\begin{tcolorbox}[
    colback = lightgray,
    colframe = darkgray,
    width   = \columnwidth,
    boxrule = 0.4pt,
    arc     = 3pt,
    left=6pt,right=6pt,top=6pt,bottom=6pt]
\textbf{Example Workflow:}
\begin{Verbatim}[fontsize=\scriptsize]

1. Information Gathering Agent: "Action: Physical Examination
   Action Input: Abdominal tenderness, McBurney's point"
   
2. RetrieveResults Tool: Returns physical exam findings
   
3. Information Gathering Agent: "Action: Laboratory Tests  
   Action Input: Complete blood count, C-reactive protein"
   
4. RetrieveResults Tool: Returns lab values
   
5. Information Interpretation Agent: Processes lab results as 
   {"WBC": {"value": 15000, "interpretation": "high"}}
   
6. Information Gathering Agent: "Action: done"
   
7. Differential Diagnosis Agent: Generates final ranked diagnosis

\end{Verbatim}
\end{tcolorbox}
\caption{Sequential processing of a patient case demonstrating the flow from information gathering through data retrieval, interpretation, and final diagnosis for acute abdominal pain.}
\end{figure}

The system incorporated robust error handling, including format validation with single retry attempts for malformed outputs and 60-second API timeouts with exponential backoff for rate limiting. When requested tests were unavailable, the workflow continued with accessible data rather than terminating. Token usage was tracked separately for each agent phase to enable precise cost calculations across heterogeneous multi-agent configurations.

\subsection{Test Set Results}
\label{sec:test}

\begin{table}[h!]
\centering
\footnotesize
\begin{tabular}{|l|c|c|c|c|c|c|c|c|}
\hline
\textbf{Agent System} & \textbf{Micro Avg} & \textbf{Macro Avg} & \textbf{Appendicitis} & \textbf{Pancreatitis} & \textbf{Cholecystitis} & \textbf{Diverticulitis} & \textbf{Top3} & \textbf{Top5} \\
 & \textbf{Accuracy} & \textbf{Accuracy} & \textbf{Accuracy} & \textbf{Accuracy} & \textbf{Accuracy} & \textbf{Accuracy} & \textbf{Accuracy} & \textbf{Accuracy} \\
\hline
multi\_gemini-flash\_gemini-flash\_gpt & 77.39 & 73.75 & 93.58 & 67.64 & 68.45 & 65.32 & 86.05 & 88.57 \\
\hline
multi\_gemini-flash\_gpt\_gpt & 77.31 & 74.69 & 93.38 & 66.55 & 66.25 & 72.58 & 87.48 & 88.91 \\
\hline
multi\_o3-mini\_o3-mini\_o3-mini & 76.97 & 73.69 & 93.8 & 64.36 & 68.04 & 68.55 & 85.88 & 88.49 \\
\hline
multi\_claude\_gpt\_gpt & 76.22 & 72.94 & 92.95 & 69.09 & 62.78 & 66.94 & 85.13 & 87.48 \\
\hline
multi\_gemini-flash\_gemini-flash\_gemini-flash & 75.88 & 71.98 & 93.63 & 57.76 & 68.77 & 67.74 & 84.79 & 87.98 \\
\hline
single\_gemini-flash\_gpt & 75.63 & 71.71 & 89.62 & 63.18 & 70.35 & 63.71 & 83.87 & 87.65 \\
\hline
multi\_claude\_claude\_gpt & 75.63 & 72.8 & 92.52 & 65.09 & 63.41 & 70.16 & 85.55 & 87.39 \\
\hline
multi\_llama\_gpt\_gpt & 75.38 & 72.87 & 88.89 & 65.09 & 68.14 & 69.35 & 85.38 & 88.32 \\
\hline
multi\_gemini\_gemini\_gemini & 74.96 & 71.53 & 87.23 & 73.55 & 64.67 & 60.66 & 85.21 & 88.91 \\
\hline
multi\_gemini\_gpt\_gpt & 74.79 & 72.01 & 88.44 & 72.73 & 62.66 & 64.23 & 84.29 & 87.73 \\
\hline
multi\_gemini\_gemini\_gpt & 74.71 & 72.09 & 88.46 & 73.45 & 60.88 & 65.57 & 84.87 & 87.14 \\
\hline
single\_gpt-4.1\_gpt & 74.37 & 71.08 & 89.1 & 71.74 & 60.57 & 62.9 & 83.19 & 85.55 \\
\hline
multi\_claude\_claude\_claude & 74.29 & 70.89 & 91.53 & 64.98 & 59.31 & 67.74 & 86.55 & 88.74 \\
\hline
single\_gpt\_gpt & 73.95 & 71.01 & 87.82 & 73.19 & 59.31 & 63.71 & 82.61 & 84.62 \\
\hline
single\_llama\_gpt & 73.95 & 70.54 & 87.82 & 66.3 & 65.93 & 62.1 & 84.79 & 87.98 \\
\hline
single\_deepseek\_gpt & 73.95 & 70.99 & 85.68 & 72.46 & 63.72 & 62.1 & 84.37 & 86.13 \\
\hline
multi\_llama\_llama\_gpt & 73.7 & 70.6 & 88.22 & 67.39 & 63.09 & 63.71 & 84.96 & 87.56 \\
\hline
multi\_claude\_gpt-4.1\_gemini-flash & 73.19 & 69.25 & 91.74 & 58.12 & 61.83 & 65.32 & 81.43 & 84.37 \\
\hline
multi\_llama\_gpt-4.1\_gemini-flash & 72.77 & 68.54 & 87.08 & 61.01 & 67.19 & 58.87 & 82.18 & 86.13 \\
\hline
multi\_llama\_llama\_llama & 72.44 & 69.98 & 85.47 & 62.18 & 66.14 & 66.13 & 84.2 & 88.07 \\
\hline
multi\_gpt\_gpt\_claude & 71.93 & 68.35 & 87.08 & 64.62 & 59.62 & 62.1 & 83.36 & 86.89 \\
\hline
multi\_gpt\_gpt-4.1\_llama & 71.85 & 69.06 & 85.04 & 62.18 & 65.3 & 63.71 & 82.52 & 86.3 \\
\hline
multi\_gpt-4.1\_gpt-4.1\_gpt-4.1 & 71.76 & 68.93 & 85.87 & 69.09 & 58.68 & 62.1 & 82.94 & 85.97 \\
\hline
single\_gemini\_gpt & 71.09 & 67.21 & 81.36 & 77.98 & 57.1 & 52.42 & 83.03 & 88.32 \\
\hline
multi\_gpt\_gpt\_gpt & 71.01 & 68.36 & 85.47 & 58.91 & 63.72 & 65.32 & 81.18 & 84.29 \\
\hline
multi\_gpt\_claude\_claude & 70.59 & 66.8 & 86.65 & 61.37 & 58.68 & 60.48 & 82.77 & 86.55 \\
\hline
multi\_gpt\_gpt-4.1\_claude & 70.59 & 67.12 & 86.44 & 59.21 & 59.94 & 62.9 & 83.19 & 87.23 \\
\hline
single\_claude\_gpt & 70.59 & 67.34 & 85.17 & 64.98 & 57.1 & 62.1 & 83.11 & 86.97 \\
\hline
multi\_gpt-4.1\_gpt\_gpt & 69.33 & 67.16 & 83.76 & 67.64 & 53 & 64.23 & 80.34 & 83.03 \\
\hline
multi\_gpt-4.1\_gpt-4.1\_gpt & 69.24 & 66.9 & 81.84 & 70.55 & 53.94 & 61.29 & 80.08 & 83.19 \\
\hline
multi\_gpt\_gpt-4.1\_gemini-flash & 67.65 & 64.6 & 81.14 & 54.15 & 61.83 & 61.29 & 76.05 & 79.16 \\
\hline
multi\_gpt\_claude\_gemini-flash & 67.39 & 63.74 & 80.08 & 53.43 & 64.98 & 56.45 & 75.97 & 79.16 \\
\hline
single\_o3-mini\_gpt & 63.45 & 58.76 & 77.78 & 57.25 & 56.47 & 43.55 & 74.12 & 78.57 \\
\hline
multi\_deepseek\_deepseek\_deepseek & 53.95 & 55.96 & 64.3 & 57.92 & 51.18 & 50.43 & 61.76 & 63.61 \\
\hline
\end{tabular}
\caption{Outcome metrics for the single and multi-agent systems on the test set}
\label{tab:model_performance}
\end{table}

\begin{table}[h!]
\centering
\scriptsize
\begin{tabular}{|l|c|c|c|c|c|c|c|c|c|}
\hline
\textbf{Agent System} & \textbf{Physical Exam} & \textbf{Physical Exam} & \textbf{Avg} & \textbf{Avg} & \textbf{Avg} & \textbf{Avg} & \textbf{Coverage} & \textbf{Coverage-} & \textbf{Lab} \\
 & \textbf{First} & \textbf{Any} & \textbf{Tools} & \textbf{Labs} & \textbf{Img} & \textbf{Physical Exam} & \textbf{} & \textbf{Test Ratio} & \textbf{Interp} \\
\hline
multi\_gemini-flash\_gemini-flash\_gpt & 72.27 & 78.15 & 6.47 & 4.25 & 1.33 & 3.43 & 0.71 & 0.11 & 82.81 \\
\hline
multi\_gemini-flash\_gpt\_gpt & 71.09 & 77.56 & 6.40 & 4.22 & 1.37 & 3.32 & 0.71 & 0.11 & 85.78 \\
\hline
multi\_o3-mini\_o3-mini\_o3-mini & 59.41 & 80.76 & 5.39 & 3.01 & 1.56 & 2.81 & 0.65 & 0.12 & 85.76 \\
\hline
multi\_claude\_gpt\_gpt & 89.16 & 94.20 & 8.27 & 5.98 & 1.35 & 6.23 & 0.80 & 0.10 & 85.43 \\
\hline
multi\_gemini-flash\_gemini-flash\_gemini-flash & 72.27 & 78.49 & 6.44 & 4.23 & 1.31 & 3.46 & 0.71 & 0.11 & 81.68 \\
\hline
single\_gemini-flash\_gpt & 49.75 & 58.15 & 7.13 & 5.23 & 1.28 & 2.46 & 0.67 & 0.09 & 80.67 \\
\hline
multi\_claude\_claude\_gpt & 88.99 & 93.95 & 8.18 & 5.88 & 1.36 & 6.18 & 0.79 & 0.10 & 85.47 \\
\hline
multi\_llama\_gpt\_gpt & 50.17 & 73.78 & 8.59 & 5.83 & 2.02 & 2.98 & 0.81 & 0.09 & 87.07 \\
\hline
multi\_gemini\_gemini\_gemini & 44.12 & 46.64 & 7.15 & 5.60 & 1.08 & 3.59 & 0.62 & 0.09 & 77.76 \\
\hline
multi\_gemini\_gpt\_gpt & 44.71 & 46.97 & 7.20 & 5.67 & 1.05 & 3.57 & 0.62 & 0.09 & 85.34 \\
\hline
multi\_gemini\_gemini\_gpt & 43.95 & 45.97 & 7.09 & 5.58 & 1.04 & 3.51 & 0.63 & 0.09 & 78.17 \\
\hline
single\_gpt-4.1\_gpt & 63.70 & 64.03 & 7.06 & 5.39 & 1.00 & 4.73 & 0.59 & 0.08 & 83.99 \\
\hline
multi\_claude\_claude\_claude & 88.91 & 94.79 & 8.25 & 5.98 & 1.33 & 6.27 & 0.79 & 0.10 & 85.22 \\
\hline
single\_gpt\_gpt & 49.92 & 50.08 & 4.25 & 2.98 & 1.03 & 0.95 & 0.52 & 0.12 & 83.36 \\
\hline
single\_llama\_gpt & 16.55 & 77.90 & 9.39 & 6.67 & 1.92 & 2.65 & 0.82 & 0.09 & 75.38 \\
\hline
single\_deepseek\_gpt & 51.26 & 64.71 & 6.78 & 4.36 & 1.11 & 3.50 & 0.60 & 0.09 & 72.63 \\
\hline
multi\_llama\_llama\_gpt & 50.42 & 75.46 & 8.65 & 5.90 & 1.99 & 3.03 & 0.81 & 0.09 & 84.80 \\
\hline
multi\_claude\_gpt-4.1\_gemini-flash & 88.99 & 94.71 & 8.26 & 5.96 & 1.35 & 6.23 & 0.80 & 0.10 & 83.94 \\
\hline
multi\_llama\_gpt-4.1\_gemini-flash & 50.59 & 75.71 & 8.97 & 6.13 & 2.04 & 3.26 & 0.82 & 0.09 & 85.78 \\
\hline
multi\_llama\_llama\_llama & 54.87 & 80.08 & 8.84 & 6.02 & 2.01 & 3.24 & 0.83 & 0.09 & 84.98 \\
\hline
multi\_gpt\_gpt\_claude & 52.35 & 52.77 & 3.61 & 1.69 & 1.34 & 2.36 & 0.46 & 0.13 & 85.50 \\
\hline
multi\_gpt\_gpt-4.1\_llama & 52.69 & 52.77 & 3.51 & 1.66 & 1.32 & 2.30 & 0.46 & 0.13 & 85.40 \\
\hline
multi\_gpt-4.1\_gpt-4.1\_gpt-4.1 & 87.82 & 88.40 & 6.91 & 5.03 & 0.96 & 6.88 & 0.59 & 0.09 & 83.54 \\
\hline
gemini\_claude & 29.66 & 30.34 & 6.29 & 5.37 & 0.61 & 2.27 & 0.53 & 0.08 & 79.23 \\
\hline
multi\_gpt\_gpt\_gpt & 52.52 & 52.94 & 3.49 & 1.66 & 1.29 & 2.35 & 0.46 & 0.13 & 84.99 \\
\hline
multi\_gpt\_claude\_claude & 50.84 & 51.01 & 3.55 & 1.70 & 1.29 & 2.33 & 0.46 & 0.13 & 82.64 \\
\hline
multi\_gpt\_gpt-4.1\_claude & 51.26 & 51.68 & 3.29 & 1.60 & 1.13 & 2.29 & 0.45 & 0.14 & 86.28 \\
\hline
single\_claude\_gpt & 69.16 & 74.37 & 7.19 & 5.52 & 0.93 & 4.62 & 0.69 & 0.10 & 82.75 \\
\hline
multi\_gpt-4.1\_gpt\_gpt & 87.90 & 88.40 & 6.77 & 4.96 & 0.90 & 6.74 & 0.60 & 0.09 & 85.71 \\
\hline
multi\_gpt-4.1\_gpt-4.1\_gpt & 88.66 & 88.74 & 6.74 & 4.88 & 0.93 & 6.96 & 0.59 & 0.09 & 83.40 \\
\hline
multi\_gpt\_gpt-4.1\_gemini-flash & 53.70 & 54.03 & 3.62 & 1.70 & 1.32 & 2.47 & 0.46 & 0.13 & 85.50 \\
\hline
multi\_gpt\_claude\_gemini-flash & 52.10 & 52.35 & 3.41 & 1.59 & 1.25 & 2.34 & 0.45 & 0.13 & 82.71 \\
\hline
single\_o3-mini\_gpt & 1.93 & 1.93 & 0.03 & 0.01 & 0.00 & 0.03 & 0.00 & 0.03 & 73.24 \\
\hline
multi\_deepseek\_deepseek\_deepseek & 81.76 & 85.88 & 7.56 & 4.86 & 1.45 & 4.96 & 0.76 & 0.10 & 82.76 \\
\hline
\end{tabular}
\caption{Process metrics for the single and multi-agent systems on the test set}
\label{tab:medical_model_performance}
\end{table}

\begin{table}[h!]
\centering
\begin{tabular}{|l|c|c|}
\hline
\textbf{Agent System} & \textbf{Computational Cost} & \textbf{Lab Cost} \\
\hline
multi\_gemini-flash\_gemini-flash\_gpt & 16.26 & 49.56 \\
\hline
multi\_gemini-flash\_gpt\_gpt & 21.29 & 53.63 \\
\hline
multi\_o3-mini\_o3-mini\_o3-mini & 14.99 & 43.43 \\
\hline
multi\_claude\_gpt\_gpt & 37.03 & 86.43 \\
\hline
multi\_gemini-flash\_gemini-flash\_gemini-flash & 12.72 & 48.38 \\
\hline
single\_gemini-flash\_gpt & 20.53 & 62.81 \\
\hline
multi\_claude\_claude\_gpt & 35.78 & 87.04 \\
\hline
multi\_llama\_gpt\_gpt & 20.89 & 58.62 \\
\hline
multi\_gemini\_gemini\_gemini & 19.62 & 80.14 \\
\hline
multi\_gemini\_gpt\_gpt & 23.34 & 99.05 \\
\hline
multi\_gemini\_gemini\_gpt & 21.20 & 94.38 \\
\hline
single\_gpt-4.1\_gpt & 42.21 & 115.76 \\
\hline
multi\_claude\_claude\_claude & 38.25 & 86.97 \\
\hline
single\_gpt\_gpt & 36.45 & 21.66 \\
\hline
single\_llama\_gpt & 26.12 & 71.71 \\
\hline
single\_deepseek\_gpt & 40.51 & 62.74 \\
\hline
multi\_llama\_llama\_gpt & 15.77 & 61.74 \\
\hline
multi\_claude\_gpt-4.1\_gemini-flash & 34.56 & 84.35 \\
\hline
multi\_llama\_gpt-4.1\_gemini-flash & 22.90 & 69.52 \\
\hline
multi\_llama\_llama\_llama & 12.06 & 57.29 \\
\hline
multi\_gpt\_gpt\_claude & 18.85 & 14.14 \\
\hline
multi\_gpt\_gpt-4.1\_llama & 15.62 & 13.43 \\
\hline
multi\_gpt-4.1\_gpt-4.1\_gpt-4.1 & 25.20 & 90.55 \\
\hline
single\_gemini\_gpt & 27.19 & 106.11 \\
\hline
multi\_gpt\_gpt\_gpt & 16.40 & 14.09 \\
\hline
multi\_gpt\_claude\_claude & 18.47 & 14.64 \\
\hline
multi\_gpt\_gpt-4.1\_claude & 20.21 & 13.78 \\
\hline
single\_claude\_gpt & 57.37 & 85.21 \\
\hline
multi\_gpt-4.1\_gpt\_gpt & 24.47 & 78.27 \\
\hline
multi\_gpt-4.1\_gpt-4.1\_gpt & 24.06 & 77.02 \\
\hline
multi\_gpt\_gpt-4.1\_gemini-flash & 15.75 & 14.36 \\
\hline
multi\_gpt\_claude\_gemini-flash & 15.22 & 14.91 \\
\hline
single\_o3-mini\_gpt & 0.61 & 0.05 \\
\hline
multi\_deepseek\_deepseek\_deepseek & 18.52 & 44.65 \\
\hline
\end{tabular}
\caption{Cost Efficiency metrics for the single and multi-agent systems on the test set}
\label{tab:cost_analysis}
\end{table}

\end{document}